\documentclass[10pt,twocolumn,letterpaper]{article}

\usepackage[pagenumbers]{iccv} % To force page numbers, e.g. for an arXiv version

\definecolor{iccvblue}{rgb}{0.21,0.49,0.74}
\usepackage[pagebackref,breaklinks,colorlinks,allcolors=iccvblue]{hyperref}

\usepackage{multirow}
\usepackage{colortbl}
\usepackage{hhline}
\usepackage{booktabs} 
\usepackage{arydshln}
\usepackage{pifont}
\usepackage{lipsum}      % 示例文本，可省略
\usepackage{graphicx}    % 插图
\usepackage{cuted}       % 用于双栏文档中插入跨栏内容
\usepackage{capt-of}     % 用于非浮动图的 caption 支持
\usepackage{float}       % 用于 H 位置参数（可选）

\def\paperID{10503} % *** Enter the Paper ID here
\def\confName{ICCV}
\def\confYear{2025}

\title{AssistPDA: An Online Video Surveillance Assistant for Video Anomaly Prediction, Detection, and Analysis}

\makeatletter
\def\thanks#1{\protected@xdef\@thanks{\@thanks
        \protect\footnotetext{#1}}}
\makeatother
\author{
Zhiwei Yang\textsuperscript{1}, \quad
Chen Gao\textsuperscript{2}, \quad
Jing Liu\textsuperscript{1\dag}\thanks{\dag Corresponding authors}, \quad
Peng Wu\textsuperscript{3}, \quad
Guansong Pang\textsuperscript{4}, \quad
Mike Zheng Shou\textsuperscript{2\dag} \\
\textsuperscript{1}Xidian University \quad
\textsuperscript{2}Show Lab, National University of Singapore \\
\textsuperscript{3}Northwestern Polytechnical University \quad
\textsuperscript{4}Singapore Management University
}

\begin{document}
\maketitle

\begin{figure*}[t]
  \centering
   \includegraphics[width=1\linewidth]{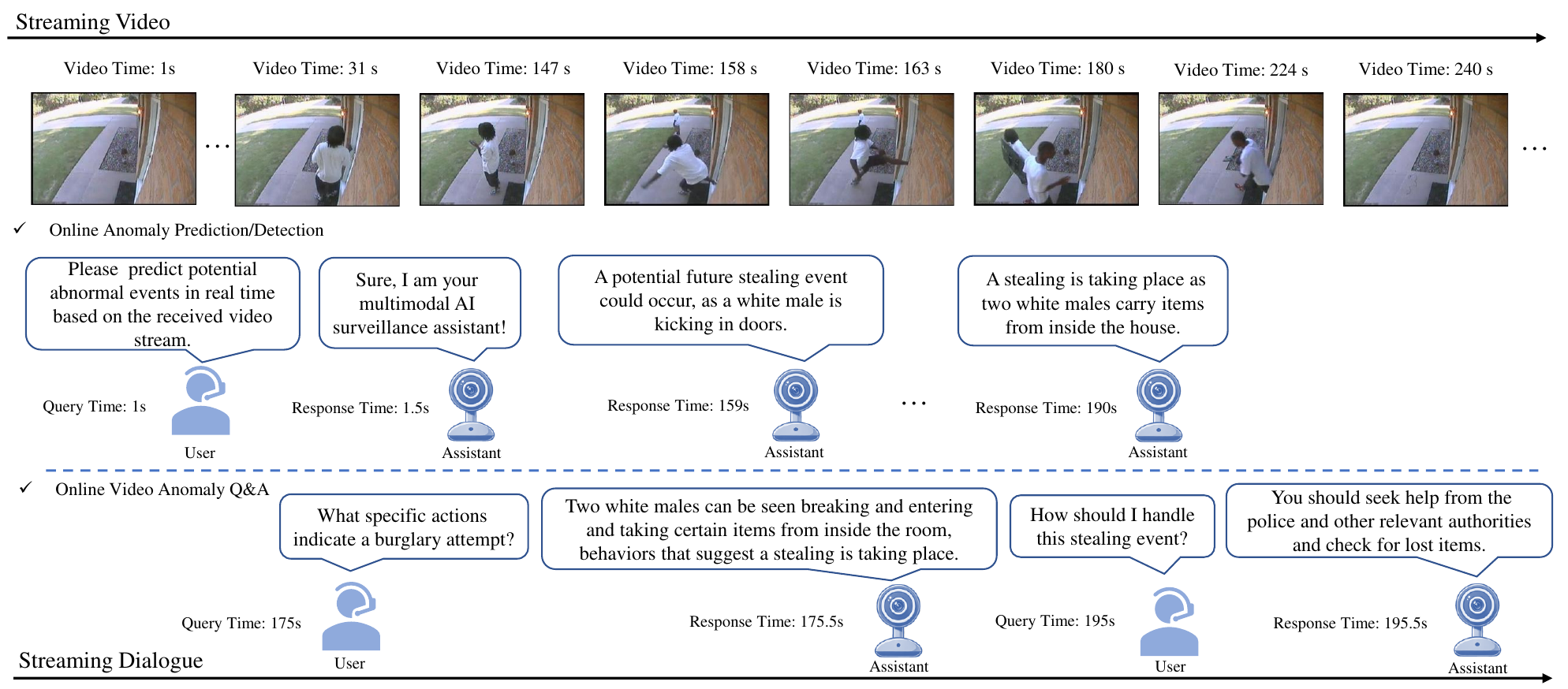}
   \vspace{-4mm}
   \caption{Illustration of the proposed Video Anomaly Prediction, Detection, and Analysis (VAPDA) tasks.}
   \vspace{-3mm}
   \label{fig: Illustration}
\end{figure*}
\begin{abstract}
The rapid advancements in large language models (LLMs) have spurred growing interest in LLM-based video anomaly detection (VAD). However, existing approaches predominantly focus on video-level anomaly question answering or offline detection, ignoring the real-time nature essential for practical VAD applications. 
To bridge this gap and facilitate the practical deployment of LLM-based VAD, we introduce \textbf{AssistPDA}, the first online video anomaly surveillance assistant that unifies video anomaly prediction, detection, and analysis (VAPDA) within a single framework. AssistPDA enables real-time inference on streaming videos while supporting interactive user engagement. 
Notably, we introduce a novel event-level anomaly prediction task, enabling proactive anomaly forecasting before anomalies fully unfold. 
To enhance the ability to model intricate spatiotemporal relationships in anomaly events, we propose a Spatio-Temporal Relation Distillation (STRD) module. STRD transfers the long-term spatiotemporal modeling capabilities of vision-language models (VLMs) from offline settings to real-time scenarios. Thus it equips AssistPDA with a robust understanding of complex temporal dependencies and long-sequence memory. 
Additionally, we construct VAPDA-127K, the first large-scale benchmark designed for VLM-based online VAPDA. Extensive experiments demonstrate that AssistPDA outperforms existing offline VLM-based approaches, setting a new state-of-the-art for real-time VAPDA. Our dataset and code will be open-sourced to facilitate further research in the community.

\end{abstract}
    
\section{Introduction}
\label{sec:intro}
Video anomaly detection (VAD)~\cite{wu2024deep-8, benezeth2009abnormal-9, zhai2016deep-10, sultani2018real-22} aims to automatically identify anomalous events in video. Traditional VAD methods mainly focus on score-based detection, \ie, assigning anomaly scores to frames, clips, or entire videos to indicate the degree of abnormality. 
However, these methods lack semantic interpretability, making them insufficient for handling complex and diverse anomalous events.

The emergence of large language models (LLMs)~\cite{touvron2023open38, zhang2023llama39, chiang2023vicuna40, liu2023visual41} has inspired LLM-based VAD approaches. For instance, Du et al. proposed an anomaly causal understanding framework~\cite{du2024uncovering-29}; Zhang et al. introduced a multimodal LLM-based unbiased and interpretable VAD framework~\cite{zhang2024holmes-07}; and Tang et al. developed an open-world anomaly comprehension method using vision-language models (VLMs)~\cite{tang2024hawk-30}. These works demonstrate the potential of LLMs in VAD, showcasing promising applications of VLMs in the field. 
However, a major limitation of these methods is that they operate in an offline setting, which fundamentally diverges from the real-world requirement for online VAD in practical surveillance scenarios. As of now, research on leveraging VLMs for online VAD remains unexplored.

% Beyond the VAD domain, online video assistants have been introduced, yet they primarily function from a first-person perspective, assisting users in answering queries related to their current activities. This paradigm differs significantly from third-person surveillance scenarios, where the system must autonomously monitor, detect, and respond to anomalies. As of now, research on online video anomaly detection leveraging large models remains an uncharted territory.

To advance the practical application of VLMs in VAD, our primary goal is to develop an online video anomaly surveillance assistant. 
% What functionalities should this video anomaly surveillance assistant possess? Answering this question serves as the motivation for our work.
Specifically, as illustrated in Fig. \ref{fig: Illustration}, we identify three core capabilities: \emph{(1) Video Anomaly Prediction (VAP):} In real-world surveillance, anomalies should not only be detected post-occurrence but also anticipated as early as possible to minimize potential damage. 
% For example, in an arson incident, early indicators such as gasoline pouring or suspicious loitering should trigger language-based warnings, alerting users to potential future anomalies. 
\emph{(2) Video Anomaly Detection:} The system must robustly detect sudden, unpredictable anomalies such as explosions or sudden attacks, ensuring timely alerts. \emph{(3) Video Anomaly Analysis (VAA):} Given the diversity of real-world anomalies, users may require real-time assistance to analyze and respond appropriately to incidents. The surveillance assistant should facilitate real-time question answering and event analysis, aiding users in handling anomalies effectively.

To realize aforementioned goals, we propose AssistPDA, the first online video surveillance assistant for Video Anomaly Prediction, Detection, and Analysis (VAPDA). AssistPDA is the first framework to unify anomaly prediction, detection, and interactive analysis within a single system, supporting real-time streaming inference and interaction. AssistPDA operates in three primary modes: proactive anomaly prediction, real-time anomaly detection, and interactive analysis. 
In predictive and detection modes, the system autonomously alerts users to critical anomalies. In interactive mode, it responds to user queries in real-time.

Developing the AssistPDA presents two key challenges. \emph{(1) Constructing training data for online VAPDA.} Existing LLM-based VAD methods have released video anomaly question-answering datasets. However, these datasets are constrained to clip-level Q\&A, making them unsuitable for training a real-time video streaming-based model. To bridge this gap, we construct VAPDA-127K, the first large-scale benchmark dataset for online VAPDA. Built upon UCF-Crime \cite{sultani2018real-22} and XD-Violence \cite{wu2020not-24} video anomaly datasets, our dataset consists of 2,415 videos across 15 anomaly categories, and 127K time-stamped anomaly predictions, detections, and Q\&A in natural language form. 
\emph{(2) Enabling temporal awareness in frame-by-frame streaming inference.} AssistPDA leverages Qwen2-VL~\cite{wang2024qwen2-06} as the backbone, which inherently supports offline video/image inference. However, transitioning to frame-by-frame online inference introduces a critical challenge since capturing long-range temporal dependencies is crucial for detecting complex and varied anomaly events.

To address this, we propose a SpatioTemporal Relation Distillation (STRD) module, inspired by recent advances in vision-language modeling. Many existing VLMs \cite{wang2024qwen2-06, team2024internvl2-05}, trained on large-scale video datasets, exhibit strong offline temporal reasoning capabilities. We aim to distill this spatiotemporal reasoning knowledge from a pre-trained offline VLM vision encoder into a lightweight module, integrating it within the online vision encoder-LLM pipeline. This enables AssistPDA to maintain robust long-term spatiotemporal understanding despite operating in a streaming frame-by-frame inference setting. Through extensive experiments, we demonstrate that AssistPDA significantly outperforms existing VLMs in VAPDA, marking a major step towards intelligent real-time video anomaly surveillance systems.
To summarize, our major contributions are as follows:
\begin{itemize}
\item We propose, for the first time, a unified framework that integrates video anomaly prediction, detection, and analysis in an online setting. Moreover, we propose event-level video anomaly prediction as a new task.

\item We devise the AssistPDA, an assistant for online video anomaly surveillance, incorporating a novel STRD module to transfer offline VLM spatiotemporal reasoning capabilities to streaming inference, significantly enhancing long-term spatiotemporal understanding.

\item We construct VAPDA-127K, the first large-scale benchmark dataset for online VAPDA, containing 127K timestamped anomaly predictions, detections, and Q\&A in natural language form, providing a valuable resource for future VLM-based video anomaly research.
\end{itemize}

\section{Related Works}
\label{sec:formatting}
\begin{figure*}[t]
  \centering
   \includegraphics[width=1\linewidth]{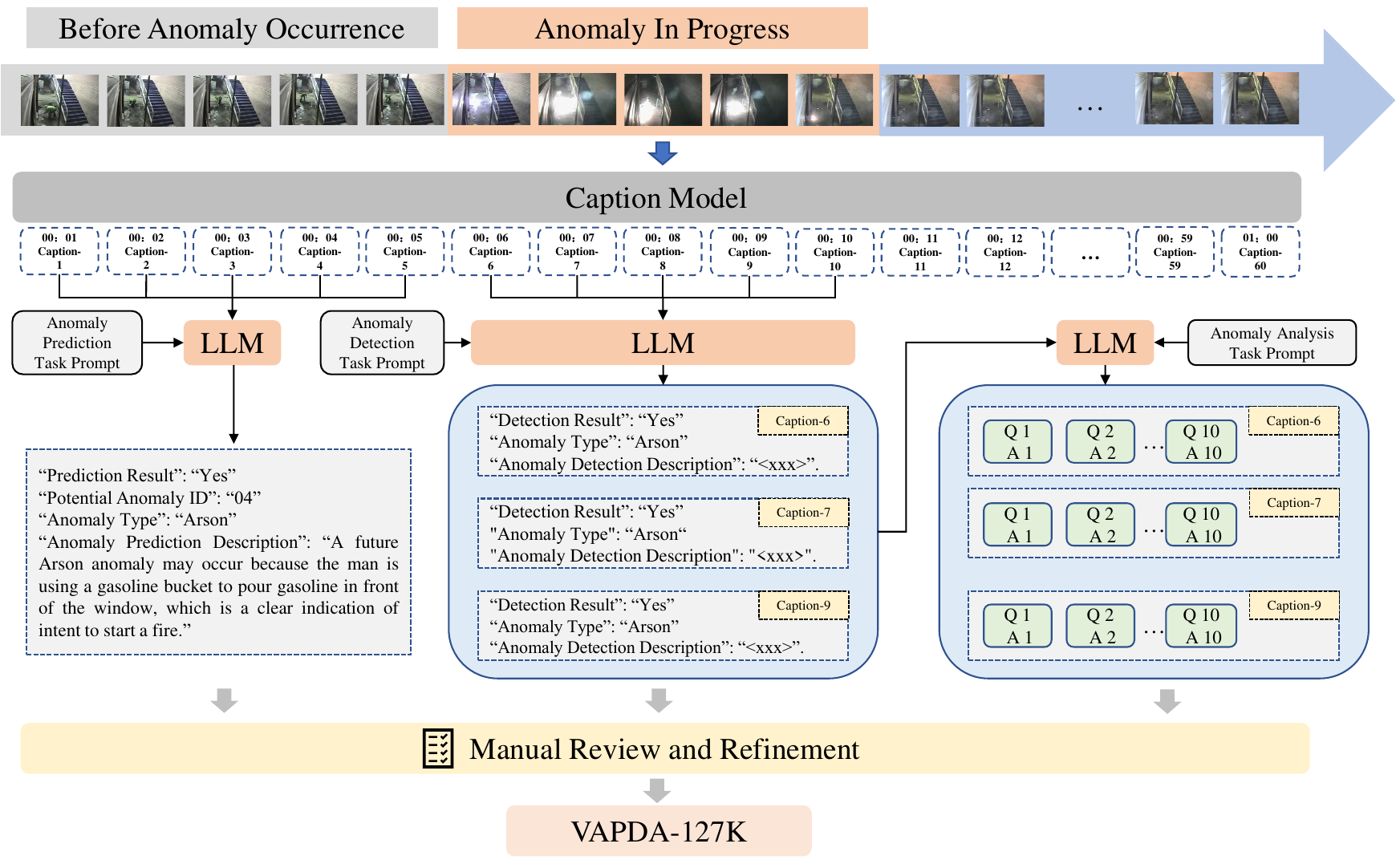}
   \caption{Pipeline of data construction for the proposed VAPDA-127K dataset.}
   \label{fig:data_construct}
\end{figure*}
\subsection{Video Anomaly Detection}
VAD problem has been studied over the years \cite{hasan2016learning-12, gong2019memorizing-11}. Early methods primarily relied on handcrafted feature-based approaches, such as those proposed in \cite{mahadevan2010anomaly-19, lu2013abnormal-18, cong2011sparse-20}. With the rapid advancements in deep learning, deep learning-based methods have become the dominant paradigm. These methods can be broadly categorized into three types: unsupervised, semi-supervised, and weakly supervised VAD.

Unsupervised VAD methods \cite{pang2020self-16} typically leverage clustering techniques or pseudo-label generation with self-training to directly mine anomaly-related information from mixed normal and abnormal data. Semi-supervised VAD approaches \cite{liu2018future-13, yang2022dynamic-15, yang2023video-17, luo2017remembering-14} are mainly based on either frame reconstruction or frame prediction, both of which employ a surrogate task to learn patterns from normal video data. During inference, deviations from normal patterns are considered anomalies. Weakly supervised VAD methods \cite{yang2024text-21, sultani2018real-22, tian2021weakly-23, wu2020not-24, wu2024vadclip-25, wu2024open-27} carry out on datasets with only video-level annotations and often utilize multiple-instance learning to infer segment-level anomaly scores.

\subsection{Multimodal Video Anomaly Detection}
With the rapid progress of LLMs and their superior performance in visual understanding \cite{gu2024anomalygpt42, wang2024clip43, wang2023few44}, multimodal VAD based on LLMs has gained increasing attention \cite{zanella2024harnessing-28, du2024uncovering-29}. For instance, Lv et al. proposed a video anomaly detection and explanation framework leveraging VLM \cite{lv2024video-33}. Zanella et al. introduced a training-free VAD method using LLMs \cite{zanella2024harnessing-28}. Du et al. proposed a framework that utilizes VLMs for causal reasoning in VAD \cite{du2024uncovering-29}. Tang et al. introduced HAWK, which leverages VLMs to understand open-world video anomalies \cite{tang2024hawk-30}. Zhang et al. developed Holmes-VAD for unbiased and interpretable VAD \cite{zhang2024holmes-07}.

However, existing VAD methods based on LLMs or VLMs are limited to single-task anomaly detection or video anomaly question answering. These methods operate only in offline settings without predictive capabilities. Such limitations hinder their applicability in real-world surveillance scenarios. 
In contrast to previous methods, we propose the first VLM-based online video anomaly surveillance assistant, unifying video anomaly prediction, detection, and real-time question answering within a single framework. 
Furthermore, we construct a large-scale benchmark dataset VAPDA-127K tailored for the online VAPDA task.

\section{Method}
In this section, we first define the tasks of video anomaly prediction, detection, and analysis in Sec 3.1.
Sec 3.2 introduces the construction process of the VAPDA-127K dataset. Sec 3.3 presents the detailed model architecture, while Sec 3.4 describes the training and inference procedures.

\subsection{Task Definition}
As mentioned, a video anomaly surveillance assistant should possess three key capabilities: \textbf{Video Anomaly Prediction}, \textbf{Video Anomaly Detection}, and \textbf{Video Anomaly Analysis}. 
We first define these tasks under the setting of VLM-based streaming video online inference.

\vspace{0.5mm}
\noindent\textbf{Video anomaly prediction}. In this work, we introduce the event-level VAP task for the first time. Although previous studies \cite{wang2024video-26, cao2023new} have explored frame-level anomaly score prediction, such as predicting whether anomalies will occur in future $T+n$ frames based on the previous $T$ frames, existing methods are typically limited to a very short prediction window (0-1s in advance), which limits their practical applicability in real-world scenarios. In contrast, event-level prediction aims to anticipate potential anomalous events before they fully unfold, leveraging historical video information to generate early warnings in natural language. The predicted output includes the event category and a descriptive explanation of the anticipated anomaly. 

% Unlike frame-level forecasting, the \textbf{temporal lead time} of event-level predictions is dynamically adjusted based on the characteristics of different anomalies.
We formalize this process as follows: At time $t_0$, a user issues a query, \eg, \textit{“Please predict potential abnormal events in real time based on the received video stream."} The actual anomaly occurs between $t_n\sim t_m$. Given the observed video stream between $t_0$ and $t_k (k<=n)$, if an anomaly is deemed likely, the model should automatically generate a natural language response at {$t_k$}, detailing the predicted event type and description.

% \begin{figure}[h]
%     \centering
%     \includegraphics[width=0.8\linewidth]{prediction_process.png}
%     \caption{Illustration of the event-level anomaly prediction process.}
%     \label{fig:prediction_process}
% \end{figure}

\vspace{0.5mm}
\noindent\textbf{Video Anomaly Detection}. Certain types of abrupt anomalous events are inherently unpredictable in advance. Therefore, the capability of the model to perform real-time anomaly detection is crucial. 
% Unlike traditional frame-level anomaly scoring approaches, which often suffer from excessive computational overhead and redundancy due to the high density of video frames, we design assistPDA to respond to the first occurrence of an anomaly as promptly as possible. To balance \textbf{timeliness} and \textbf{efficiency}, after the initial detection response, we introduce a \textbf{minimum time interval} for subsequent responses. This mechanism prevents excessive frame-level computations and high-frequency detections that might otherwise lead to missing critical frames.
We formalize this process as follows: At time $t_0$, the user issues a query, \eg, \textit{``Please detect any abnormal events in real time based on the received video stream.''} The actual anomaly occurs between $t_n\sim t_m$. During this period, the model provides anomaly detection responses at multiple key moments within $t_n$ to $t_m$, each response containing the detected anomaly type and a descriptive explanation of the event.

% \begin{figure}[h]
%     \centering
%     \includegraphics[width=0.8\linewidth]{anomaly_detection_process.png}
%     \caption{Illustration of the real-time anomaly detection process.}
%     \label{fig:anomaly_detection_process}
% \end{figure}

\vspace{0.5mm}
\noindent\textbf{Video Anomaly Analysis}. For the VAA task, we adopt a user-centric approach, where the model provides responses based on the user's inquiries regarding ongoing anomalous events. Since real-world anomalies can be highly diverse, and user queries are completely open-ended, we formalize video anomaly analysis as an online video question answering task. Specifically, assume that an anomaly begins to occur at $t_n$. At a later time $t_n + k$, the user issues a query, such as \textit{``How should the ongoing anomaly be handled?''} Upon receiving the query, the model should generate an immediate response at $t_n + l$ with $(l>=k)$, addressing the user's question in real time.

% This formulation enables assistPDA to provide timely and context-aware explanations, facilitating intelligent human-AI interaction in anomaly management scenarios.

\begin{figure*}[t]
  \centering
   \includegraphics[width=1\linewidth]{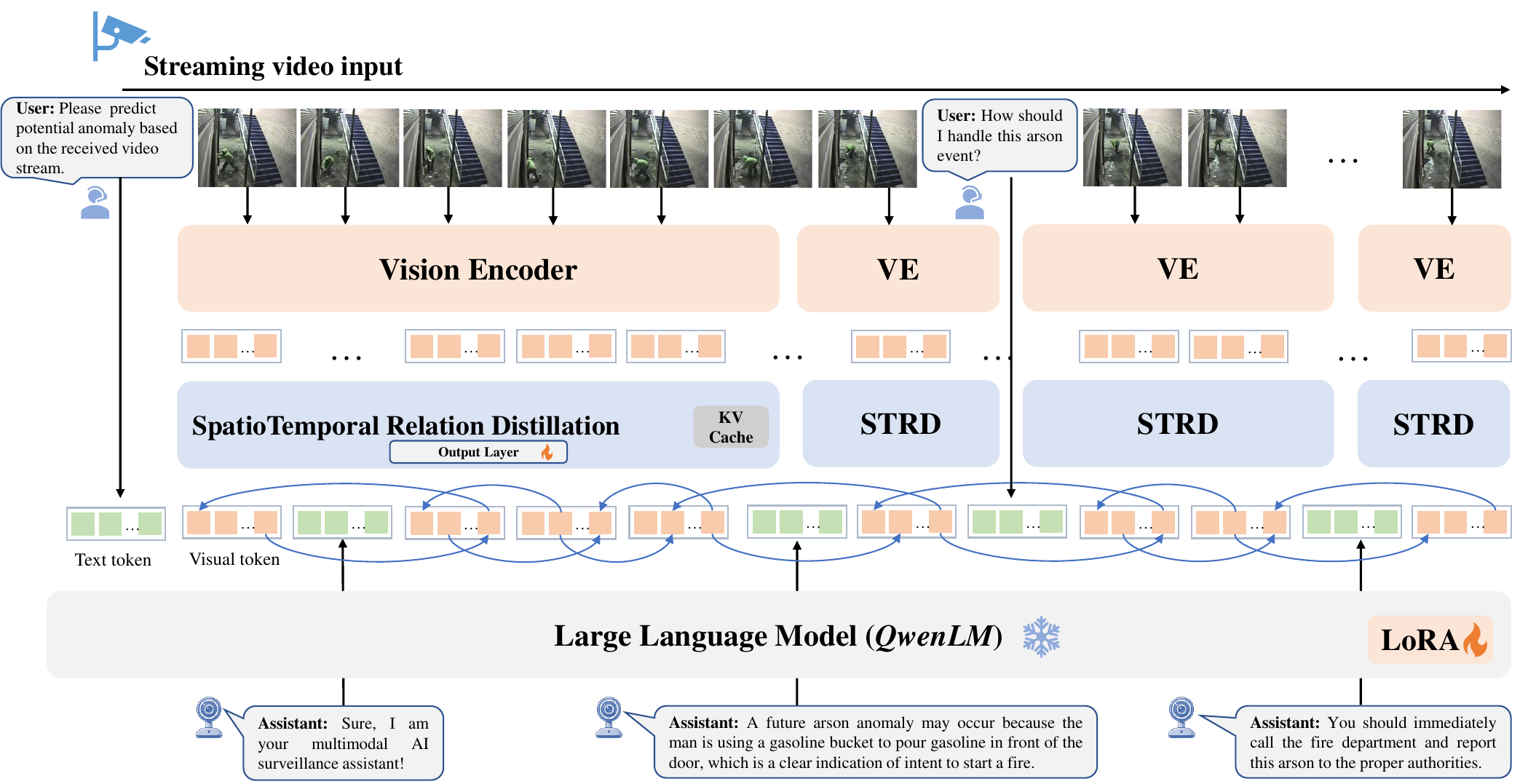}
   \vspace{-3mm}
   \caption{Pipeline of our method. VE and STRD are short for Video Encoder and Spatiotemporal relation distillation, respectively.}
   \label{fig: Pipeline}
\end{figure*}
\subsection{Dataset Construction}

In this section, we detail the construction process of the \emph{VAPDA-127K} dataset to adapt the three tasks above. Fig. \ref{fig:data_construct} illustrates the process of construction of the dataset.

% \vspace{0.5mm}
\noindent\textbf{Data Collection}. We first collect raw video data from the two largest weakly supervised VAD datasets,  UCF-Crime \cite{sultani2018real-22} and XD-Violence \cite{wu2020not-24}. After filtering out low-quality videos, we obtain a total of 2415 untrimmed videos. These videos only contain video-level annotations, indicating whether an anomaly occurs within the video, but lack precise timestamps for when anomalies happen. However, to train the VLM-online for our three defined tasks, timestamp-level anomaly annotations are necessary. Thanks to HIVAU-70K \cite{zhang2024holmes-32} providing annotated event start and end timestamps for UCF-Crime and XD-Violence, we further build our task-specific dataset based on them.

% \vspace{0.5mm}
\noindent\textbf{Data Annotation for Anomaly Prediction}. For the VAP task, we require frame-level information preceding the occurrence of an anomaly. To reduce the computational burden caused by redundant frames, we first sample the raw videos at 1 FPS and then use an existing VLM to generate a caption for each frame. Next, we segment each video at the start time of each anomalous event. We then feed all captions (with their corresponding caption ID ) from the video start time up to the onset of the anomaly into a LLM. Using specifically designed prompts, we instruct the LLM to determine the earliest frame where a potential future anomaly could have been predicted and to generate the anomaly type and a brief description of the predicted anomaly.

% \vspace{0.5mm}
\noindent\textbf{Data Annotation for Anomaly Detection}. For the VAD task, we focus on video segments corresponding to the actual anomaly occurrence. Using HIVAU-70K \cite{zhang2024holmes-32}, which provides event start and end timestamps and segment-level captions, we first extract data containing explicit start and end timestamps for anomalous events. While segment-level captions exist within the anomalous event period, not all captions within this period necessarily contain explicit anomaly-related information due to the complexity of real-world events. We sequentially feed captions with timestamps along with historical captions into the LLM. The LLM is instructed to determine whether each caption contains an ongoing anomalous event. Furthermore, by leveraging both the current and historical captions, the LLM generates a concise anomaly description that includes the anomaly type. Through this process, we obtain timestamped anomaly detection captions corresponding to key moments during the anomaly occurrence.

% \vspace{0.5mm}
\noindent\textbf{Data Annotation for Anomaly Analysis}. For the VAA task, we construct open-ended question-answer pairs based on ongoing anomalous events. This is distinctly different from existing anomalous Q\&A data, which are fixed-template Q\&A pairs constructed based on the entire video or clips. Building upon the anomaly detection annotations, we extract key detection captions at critical moments within the anomaly period and combine them with historical captions in chronological order. These are then fed into an LLM, which, based on the \emph{5W (Who, What, When, Where, Why) and 2H (How, How much)} principle, generates questions relevant to the ongoing anomalous event. The LLM also generates factually and logically consistent answers based on both the current and historical captions.

% \vspace{0.5mm}
\noindent\textbf{Manual Review and Refinement}. To mitigate the effects of LLM hallucinations, we iteratively refine the prompts to ensure optimal generated responses. Finally, all LLM-generated data undergo manual review, where inappropriate responses are removed or modified. This review process involved five annotators, each spending an average of 10 hours, ensuring high-quality annotations for the dataset. Please refer to the supplementary material for more details on the construction of the dataset.

\subsection{Model Architecture}

Fig.~\ref{fig: Pipeline} presents an overview of our proposed AssistPDA, which consists of three key components: a vision encoder, a spatiotemporal relationship distillation module, and an LLM with a fine-tunable LoRA module. In the following sections, we provide details of each module.

\subsubsection{Vision Encoder}
We adopt the frozen vision encoder ${\varphi _v}$ from Qwen2-VL \cite{wang2024qwen2-06}, which is based on a Vision Transformer (ViT)~\cite{dosovitskiy2020image45}.
% During video preprocessing, Qwen2-VL supports naive dynamic resolutions and applies a smart resizing mechanism by setting a maximum pixel value. Here, we also adopt this functionality to maximize GPU resource utilization while avoiding GPU constraints imposed by processing long videos. These constraints would otherwise force an excessively low resolution for short videos, leading to an insufficient number of vision tokens and ultimately hindering vision understanding. This limitation would force the use of overly low resolutions, resulting in an insufficient number of tokens for short videos and ultimately hindering vision understanding.
Following existing work~\cite{chen2024videollm-01}, we sample frames from the original video at $2$ FPS. To support both image and video inputs, the Qwen2-VL vision encoder duplicates input images when operating in image mode. To reduce redundant computation, we directly take every two consecutive frames as input to extract visual tokens. Given an input video frame sequence $\nu \in \mathbb{R}^{T \times H \times W \times C}$, the visual token obtained from the $(i-1)$-th and $i$-th frames is formulated as: 
\begin{equation}
\small
{V_{i - 1,i}} = \{ v_{i - 1,i}^1,\;v_{i - 1,i}^2,\;v_{i - 1,i}^3,\;...,\;v_{i - 1,i}^N\}  = {\varphi _v}({\nu_{i - 1}},{\nu_i}),
\end{equation}
where $v_{i - 1,i}^j\;(j \in \{ 1,2,\;...,\;N\} )$ represents the patch tokens, with $N$ denoting the total number of patches obtained from every two input frames. For clarity and conciseness, we will refer to ``each frame" as a representation of the actual two-frame input in the subsequent discussion.

\begin{figure}[t]
  \centering
   \includegraphics[width=1\linewidth]{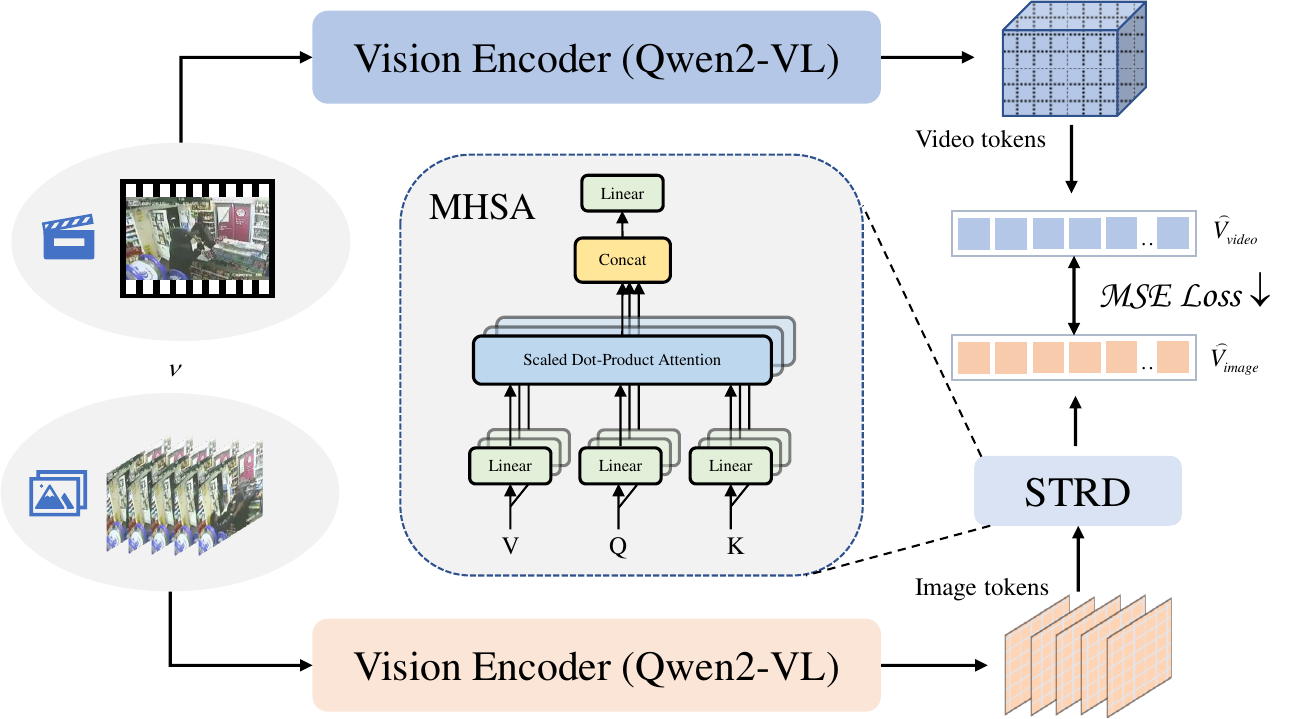}
   \caption{Illustration of the STRD module.}
   \label{fig: STRD}
\end{figure}

\subsubsection{SpatioTemporal Relationship Distillation}

In online processing mode, video frames are input frame by frame, making the learning of spatiotemporal relationships and long-term memory a significant challenge. Existing approaches often incorporate memory units between the vision encoder and the LLM to store historical frame information, which is then retrieved to maintain temporal memory or extract key information. However, such methods impose substantial constraints on inference speed.

To ensure that our designed online framework, AssistantPDA, maintains high inference efficiency while also exhibiting strong spatiotemporal reasoning and long-term memory capabilities, we introduce a STRD module ${\phi}$. To minimize additional computational overhead on the VLM backbone, we adopt a lightweight approach by employing a two-layer Multi-Head Self-Attention (MHSA) network as the STRD module. This module transfers the VLM's offline-mode ability to model global spatiotemporal relationships into an online processing pipeline. We perform the distillation using Qwen2-VL \cite{wang2024qwen2-06} with the goal that the tokens obtained from the frame-by-frame video input remain as consistent as possible in feature space with those obtained from processing the entire video directly. The spatiotemporal relationship distillation process is illustrated in Fig. \ref{fig: STRD}. First, the input video frame sequence $\nu \in \mathbb{R}^{T \times H \times W \times C}$ is directly processed by the Qwen2-VL vision encoder, obtaining the global visual token representation:
\begin{equation}
\overset{\lower0.5em\hbox{$\smash{\scriptscriptstyle\frown}$}}{V} _{video} = \{ v_i^1, v_i^2, v_i^3, ..., v_i^M\} = {\varphi _v}(\nu),
\end{equation}
where $v_i^j$ ($j \in \{1,2,...,M\}$) denotes the patch tokens and $M$ is the number of patches extracted from the input video.

Since the vision encoder applies a 3D convolution with a stride of 2 before patch embedding, each $v_i^j$ still represents a patch token fused from two consecutive frames. However, unlike frame-by-frame input, since the visual tokens here are obtained through a global attention operation, meaning each $v_i^j$ inherently contains information from all other frame patches, incorporating full spatiotemporal context. The role of the STRD module is to ensure that tokens obtained from frame-by-frame input, after transformation, also encapsulate global contextual information. To achieve this, we first concatenate the tokens obtained from frame-by-frame input along the temporal dimension:
\begin{equation}
{V_{images}} = concate(V_{i - 1,i}^1,\;V_{i + 1,i + 2}^2,\;...,\;V_{T - 1,T}^{T/2}).
\end{equation}
We then apply the distillation module to transform these tokens, which is formulated as:
\begin{equation}
\overset{\lower0.5em\hbox{$\smash{\scriptscriptstyle\frown}$}}{V} _{images}^{} = \phi (V_{images}^{}).
\end{equation}
Finally, we enforce consistency between $\overset{\lower0.5em\hbox{$\smash{\scriptscriptstyle\frown}$}}{V} _{images}^{}$ and the global video tokens $\overset{\lower0.5em\hbox{$\smash{\scriptscriptstyle\frown}$}}{V} _{video}^{}$ in feature space using a mean squared error (MSE) loss function:
\begin{equation}
{L_{distill}} = \frac{1}{M}\left\| {\overset{\lower0.5em\hbox{$\smash{\scriptscriptstyle\frown}$}}{V} _{video}^{} - \overset{\lower0.5em\hbox{$\smash{\scriptscriptstyle\frown}$}}{V} _{image}^{}} \right\|_2^2.
\end{equation}

After training the STRD module, we insert it between the vision encoder and the LLM during LoRA fine-tuning. In real-time inference, the MHSA module in the STRD module is equipped with KV cache, allowing frame-by-frame input tokens to retain historical spatiotemporal context. By adjusting the length of the KV cache, we can control the temporal span of frames considered by the STRD module. On our experimental setup with an A6000 GPU, the maximum temporal receptive field can reach up to $20$ minutes.

\subsubsection{LLM}

The LLM used in our framework is QwenLM from Qwen2-VL \cite{wang2024qwen2-06}. It is responsible for processing the visual tokens obtained from the STRD module, concatenating them with the text tokens derived from the user query in temporal order, and feeding them into the LLM for decoding to generate the VLM response.

\subsection{Training and Inference}

Our training process consists of two stages. The first stage involves pre-training the STRD module. As described in Sec 3.3.2, we optimize this module using the MSE loss function. The second stage involves instruction fine-tuning of the model using the constructed VAPDA-127K dataset. The loss function consists of two components. The first component is autoregressive language modeling, which aims to maximize the joint probability $P_i^{[{\text{Txt}}_{i + 1}]}$ of the input text sequence. The second component is video streaming input prediction modeling. For real-time anomaly prediction and detection tasks, AssistPDA needs to have the capability to respond automatically, determining when to generate a response and when to remain silent. Following the work \cite{chen2024videollm-01}, we introduce an additional streaming End-of-Sequence (EOS) token appended to each video frame token. The probability $P_i^{[{\text{EOS}}]}$ of predicting the EOS token is used to decide whether to continue receiving video frame inputs or to generate a response. Both components are optimized using the cross-entropy loss function, formulated as follows:
\begin{equation}
L = \frac{1}{N} \sum\limits_{i = 1}^N \left( - \log l_{i + 1} P_i^{[{\text{Txt}}_{i + 1}]} - w \log f_i P_i^{[{\text{EOS}}]} \right),
\end{equation}
where \( l_i \) and \( f_i \) are condition indicators; \( l_i \) is 1 if the \( i \)-th token is a language response token, and 0 otherwise; \( f_i \) is 1 if (1) the \( i \)-th token is the last token of a frame, and (2) \( l_{i+1} = 0 \).  \( w \) is balance term. Essentially, the streaming EOS loss is applied to frames before responding. \( P_i^{[\text{Txt}_{i+1}]} \) denotes the probability of the \( (i+1) \)-th text token output from the language model head at the \( i \)-th token, while \( P_i^{[\text{EOS}]} \) represents the probability assigned to the EOS token.

During the inference stage, AssistPDA executes different tasks based on user-specified queries. For VAP and VAD tasks, we introduce a threshold $\gamma$ to control the prediction of the EOS token. When the predicted probability of the EOS token falls below $\gamma$, the model generates a response, enabling AssistPDA to provide predictions or detection alerts at critical moments while remaining silent during normal periods. For anomaly analysis tasks, AssistPDA responds immediately after the user completes their query, no threshold setting is required. On an A6000 GPU, AssistPDA achieves an average inference speed of \textit{\textbf{15–20 FPS}}.

\section{Experiments}
\subsection{Dataset and Evaluation Metrics}
\vspace{0.5mm}
\noindent\textbf{Dataset.}
Our VAPDA-127K is constructed based on the raw videos from the two largest-scale VAD datasets, UCF-Crime~\cite{sultani2018real-22} and XD-Violence~\cite{wu2020not-24}. VAPDA-127K consists of $2,415$ untrimmed videos, covering a total of $15$ categories of anomalous events, including abuse, arson, car accidents, fighting, explosion, riots, stealing, and shooting, \etc. These events are from real-world scenarios and selected footage from movies and live broadcasts. More details about the dataset are in the supplementary material.

\vspace{0.5mm}
\noindent\textbf{Evaluation Metrics.}
For training and ablation studies, we follow~\cite{chen2024videollm-01} and adopt three evaluation metrics to efficiently assess the overall performance of the model: Language Modeling Perplexity (LM-PPL), Time Difference (TimeDiff), and Fluency. LM-PPL is a commonly used perplexity measure to evaluate language modeling capability, where a lower LM-PPL indicates more accurate responses. TimeDiff measures the temporal alignment ability of the model by computing the difference between the predicted response timestamp and the expected timestamp. Fluency evaluates the proportion of continuously and successfully predicted tokens within a dialogue round. Since this also includes language tokens, the fluency metric comprehensively reflects the model’s language modeling ability in an online streaming pipeline. During inference, different evaluation metrics are used depending on the task. For textual responses in real-time inference across VAP, VAD, and VAA tasks,  following the work\cite{du2024uncovering-29} we employ MoverScore (MS) \cite{zhao2019moverscore-35}, Bleurt \cite{sellam2020bleurt-34}, and Unieval \cite{zhong2022towards-unieval-36} to evaluate response quality by comparing them with ground truth text annotations. For VAP and VAD tasks, we also use the weighted F1-score to measure the model’s accuracy in classifying predicted and detected anomaly types. In addition, we introduce the average advance time (AAT) metric to evaluate the model’s capability to predict anomalies in advance by comparing the response time with the actual anomaly onset time. 

\subsection{Implementation Details}
The vision encoder and LLM module of our framework are initialized using the Qwen2-VL-2B-Instruct version. During the pretraining distillation stage of the STRD module, we train for a total of 10 epochs using the AdamW optimizer with an initial learning rate of \(1 \times 10^{-4}\), employing a cosine annealing decay strategy. In the main framework training stage, we train all linear layers of the LLM using LoRA with $r=32$ and $\alpha=64$, and the epoch is set to 2. Additionally, we fine-tune the final output linear layer of the STRD module. The default loss weight \( w \) is set to 1. The EOS token prediction thresholds for video anomaly prediction and detection are set to 0.96 and 0.7. Further execution details can be found in the supplementary materials.

\begin{table*}
\centering
\setlength{\tabcolsep}{3pt} 
% \small
% \footnotesize
\scriptsize
\arrayrulecolor{black}
\begin{tabular}{lccccccccccccc} 
\arrayrulecolor{black}\hline
\multicolumn{1}{c}{Task Name}               & ~      & \multicolumn{5}{c}{VAP}                                                                        & \multicolumn{4}{c}{VAD}                                               & \multicolumn{3}{c}{VAA}               \\ 
\arrayrulecolor{black}\cmidrule(lr){1-2}\cmidrule(lr){3-7}\cmidrule(lr){8-11}\cmidrule(lr){12-14}
\multicolumn{1}{c}{~}                       & ~      & \multirow{2}{*}{F1-score(\%)} & \multirow{2}{*}{AAT (s)} & \multicolumn{3}{c}{Language}         & \multirow{2}{*}{F1-score (\%)} & \multicolumn{3}{c}{Language}         & \multicolumn{3}{c}{Language}          \\ 
\arrayrulecolor{black}\cmidrule(lr){1-2}\cmidrule(lr){5-7}\cmidrule(lr){9-11}\cmidrule(lr){12-14}
\multicolumn{1}{c}{~}                       & Params &                               &                         & MS (\%) & Bleurt (\%) & Unieval (\%) &                                & MS (\%) & Bleurt (\%) & Unieval (\%) & MS (\%) & Bleurt (\%) & Unieval (\%)  \\ 
\arrayrulecolor{black}\hline
\multicolumn{14}{l}{{\cellcolor[rgb]{0.949,0.949,0.949}}\textbf{Sliding windows size=5s fps=2}}                                                                                                                                                                       \\ 
\hline
Video-LLaMA2 \cite{cheng2024videollama-02}                                & 7B     & 28.26                         & 10.32                   & 53.05   & 37.98       & 79.84        & 9.56                           & 50.65   & 29.67       & 65.23        & 56.64   & 52.15       & 80.22         \\
Video-LLaVA \cite{Video-llava-03}                                 & 7B     & 38.63                         & 12.34                   & 52.99   & 37.39       & 73.89        & 12.01                          & 48.32   & 20.28       & 67.69        & 57.05   & 44.49       & 81.84         \\
Video-ChatGPT \cite{Video-chatgpt-04}                               & 7B     & 18.94                         & 7.25                    & 53.54   & 38.06       & 41.08        & 11.35                          & 54.15   & 40.66       & 64.12        & 56.29   & 47.56       & 80.60         \\
InternVL2 \cite{team2024internvl2-05}                                  & 2B     & 16.16                         & 6.32                    & 53.22   & 33.98       & 61.98        & 13.77                          & 52.59   & 38.82       & 66.25        & 55.82   & 44.19       & 73.29         \\
Qwen2-VL \cite{wang2024qwen2-06}                                   & 2B     & 30.71                         & 11.64                   & 54.59   & 40.02       & 72.12        & 11.83                          & 55.45   & 48.14       & 69.29        & 54.78   & 47.13       & 75.45         \\
Holmes-VAD \cite{zhang2024holmes-07}                                 & 7B     & 47.91                         & 15.68                   & 54.97   & 41.47       & 70.61        & 25.83                          & 55.00   & 42.52       & 68.48        & 55.70   & 40.72       & 88.05         \\ 
\hline
\multicolumn{14}{l}{{\cellcolor[rgb]{0.949,0.949,0.949}}\textbf{Streaming video input at 2 fps}}                                                                                                                                                                      \\ 
\hline
VideoLLM-online \cite{chen2024videollm-01}                             & 8B     & 0                             & /                       & 5.23    & 6.75        & 10.23        & 0                              & 6.78    & 4.43        & 9.46         & 5.67    & 3.56        & 8.92          \\
\rowcolor[rgb]{0.851,0.851,0.851} AssistPDA & 2B     & \textbf{64.69}                         & \textbf{29.19}                   & \textbf{61.89}   & \textbf{51.63}       & \textbf{76.69}        & \textbf{45.66}                          & \textbf{65.45}   & \textbf{63.83}       & \textbf{72.46}        & \textbf{62.87}   & \textbf{61.12}       & \textbf{88.32}         \\
\hline
\end{tabular}
\arrayrulecolor{black}

\caption{Main results of VAP, VAD, VAA on VAPDA-127K. We compare existing general-purpose VLMs with those tailor-designed for VAD. For VLMs that do not support online inference, video is sampled at 2 FPS and processed using a 5-second sliding window as input.}
\label{Table: main_results}

\end{table*}

\subsection{Main Results}

Since AssistPDA is the first framework that leverages a VLM for online VAPDA, we compare it with two types of baselines: general-purpose VLMs and VLMs designed for VAD. Most existing VLMs only support offline processing. To facilitate a fair comparison, we simulate an online setting for offline models by adopting a sliding window approach with a window size of 5 seconds. Specifically, we provide each VLM with prompts related to prediction, detection, and question, instructing them to generate anomaly categories, corresponding anomaly descriptions, and responses to user queries based on the input video segments. Due to differences in instruction fine-tuning data across VLMs, we optimize prompts for each model to maximize response accuracy. For VLMs that support online processing, such as VideoLLM-online \cite{chen2024videollm-01}, we directly feed streaming video input at 2 FPS. Table \ref{Table: main_results} presents the comparison results of AssistPDA with existing methods on the three tasks. The compared VLMs include Video-LLaMA2 \cite{cheng2024videollama-02}, Video-LLaVA \cite{Video-llava-03}, Video-ChatGPT \cite{Video-chatgpt-04}, InternVL2 \cite{team2024internvl2-05}, and Qwen2-VL \cite{wang2024qwen2-06}, which are among the most advanced VLMs currently available. Additionally, Holmes-VAD \cite{zhang2024holmes-07} is an offline VLM specifically designed for VAD.

As shown in Table \ref{Table: main_results}, our method significantly outperforms all baselines, achieving superior performance across all evaluation metrics. Except for Holmes-VAD, other VLMs exhibit low F1-scores on both VAP and VAD tasks. This is primarily due to the fact that VAPDA-127K encompasses 15 distinct categories of anomalous events, posing a considerable challenge for general-purpose VLMs. The online method, VideoLLM-Online, fails to follow our instructions, producing largely garbled and redundant outputs, resulting in poor performance. Notably, in the video anomalous event prediction task, the average advance prediction time AAT of our method is 29.19s, which is a qualitative leap compared to frame-level prediction. The consistent superiority of our method across all metrics demonstrates the effectiveness of AssistPDA in VAP, VAD, and VAA tasks.

\begin{table}
\centering
\setlength{\tabcolsep}{1pt} 
% \small
% \footnotesize
\scriptsize
\arrayrulecolor{black}
\begin{tabular}{lcccccc} 
\hline
\multicolumn{1}{c}{~}                        & \multicolumn{3}{c}{VAP}      & \multicolumn{3}{c}{VAD}       \\ 
\cmidrule(lr){1-1}\cmidrule(lr){2-4}\cmidrule(lr){5-7}
\multicolumn{1}{c}{~}                        & LM-PPL $\downarrow$ & TimeDiff $\downarrow$ & Fluency  $\uparrow$­ & LM-PPL $\downarrow$ & TimeDiff $\downarrow$ & Fluency $\uparrow$­  \\ 
\hline
Baseline                                     & 1.76   & 1.52     & 53.02\%  & 2.15   & 5.14     & 46.69\%   \\
w/o pretraining                              & 1.79   & 1.85     & 52.76\%  & 2.27   & 5.19     & 44.53\%   \\
w/o finetune                                 & 1.70   & 1.27     & 53.42\%  & 1.98   & 4.82     & 46.68\%     \\
\rowcolor[rgb]{0.851,0.851,0.851} w finetune & \textbf{1.68}   & \textbf{1.07}     & \textbf{53.81\%} & \textbf{1.96}   & \textbf{4.71}     &\textbf{46.83\%}   \\
% \hhline{>{\arrayrulecolor{black}}-->{\arrayrulecolor{black}}-----}
\hline
\end{tabular}
\caption{Performance comparison of our method with different STRD settings.}
\label{Table: STRD_settings}
\end{table}

\begin{table}
\centering
\setlength{\tabcolsep}{1pt} 
% \small
% \footnotesize
\scriptsize
\arrayrulecolor{black}
\begin{tabular}{ccccccc} 
\hline
~ ~~                                       & \multicolumn{3}{c}{VAP}      & \multicolumn{3}{c}{VAD}       \\ 
\cmidrule(lr){1-1}\cmidrule(lr){2-4}\cmidrule(lr){5-7}
~                                              & LM-PPL $\downarrow$ & TimeDiff $\downarrow$ & Fluency $\uparrow$­ & LM-PPL $\downarrow$ & TimeDiff $\downarrow$ & Fluency $\uparrow$­  \\ 
\hline
1 layer MHSA                                   & \textbf{1.67}   & 1.08     & 53.07\%  & 1.97   & 4.78     & 45.95\%   \\
\rowcolor[rgb]{0.851,0.851,0.851} 2 layer MHSA & 1.68   & \textbf{1.07}     & \textbf{53.81\%} & \textbf{1.96}   & \textbf{4.71}     & \textbf{46.83\%}   \\
3 layer MHSA                                   & 1.70   & 1.09     & 53.46\%  & \textbf{1.96}   & 4.83     & 46.46\%   \\
\arrayrulecolor{black}\hline
\end{tabular}
\arrayrulecolor{black}
\caption{Performance comparison of our method on STRD with different numbers of MHSA layers.}
\label{Table: STRD_with_MHSA}
\end{table}

\subsection{Ablation Study}

We conduct ablation experiments in this subsection to analyze the effectiveness of each component of our framework.
% \subsubsection{}

\noindent\textbf{Effectiveness of the STRD.}
To evaluate the effectiveness of the STRD module, we perform multiple ablation and comparative experiments. Table \ref{Table: STRD_settings} presents the results of four experimental settings: (1) the baseline model without the STRD (baseline), (2) the model with the STRD module but without distillation pertaining (w/o pretraining), (3) the model with the STRD module but without finetune (w/o finetune), and (4) the model with both the STRD module and fine-tuning (w finetune). From the Table \ref{Table: STRD_settings}, it can be observed that the setting with both the STRD module and finetuning achieves the best performance. 
% In contrast, the setting without distillation training but with the STRD module performs the worst. This setting is somewhat similar to the commonly used projector module in existing methods, which aims to better align visual tokens with text tokens. However, due to the limited training data and the complexity of anomaly events, it fails to fully adapt. 
Compared to the baseline, the model with the STRD module and fine-tuning shows a significant advantage in both LM-PPL and TimeDiff metrics, indicating improved language modeling accuracy and temporal alignment. These results demonstrate that the design of the STRD module, combined with fine-tuning, effectively enhances the model’s capability in spatiotemporal reasoning. Table \ref{Table: STRD_with_MHSA} presents the impact of different numbers of MHSA layers in the STRD module. The results show that the 2-layer MHSA configuration achieves the best overall performance.
\begin{figure}[t]
  \centering
   \includegraphics[width=1\linewidth]{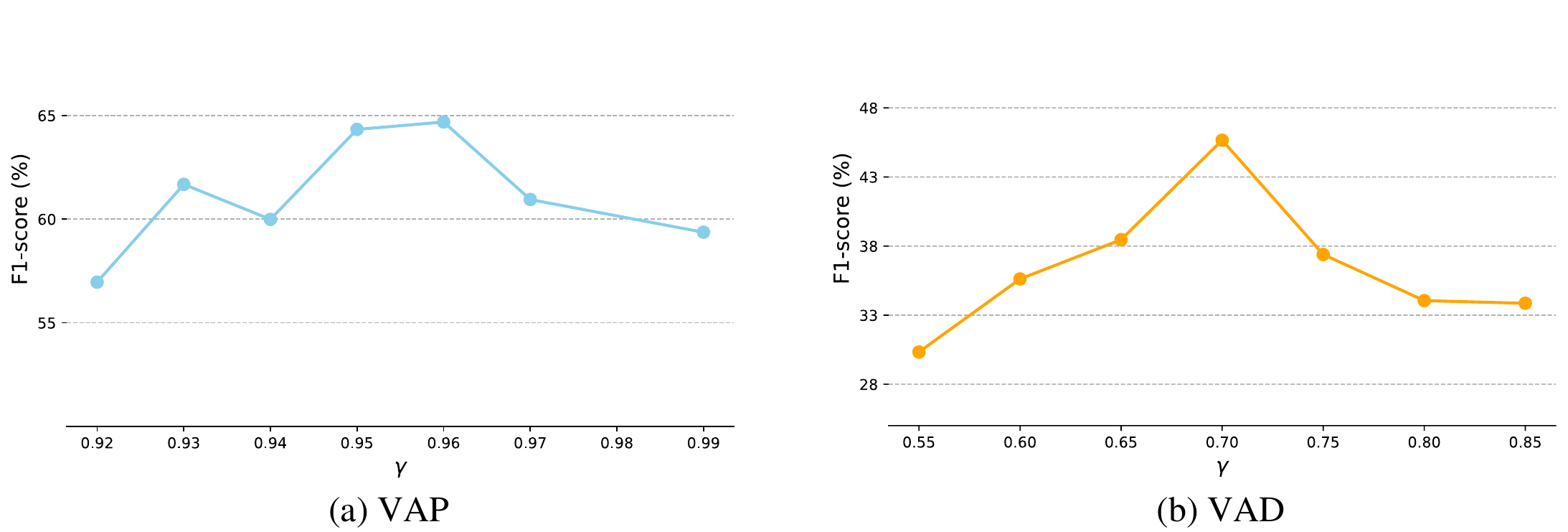}
   \caption{F1-score variation for different EOS token prediction thresholds $\gamma$ on VAP and VAD tasks.}
   \label{fig: EOS_VAP_VAD}
\end{figure}
% \subsubsection{}

\noindent\textbf{Impact of EOS Token Prediction Threshold $\gamma$.}
As shown in Fig.~\ref{fig: EOS_VAP_VAD} (a)(b), we illustrate the impact of the EOS token threshold $\gamma$ on model performance during the inference phase. 
We can observe that the prediction task is sensitive to the EOS token threshold, which aligns with its inherent nature, requiring heightened sensitivity to anomalous events. Finally, the optimal EOS token threshold $\gamma$ is set to 0.96 for the prediction task and 0.7 for the detection task.

\section{Conclusion}
In this work, we propose the AssistPDA, an online video anomaly surveillance assistant that integrates video anomaly prediction, detection, and analysis. Based on this framework, we introduce a novel event-level video anomaly prediction task aimed at enabling early warning of anomalous events. To enhance AssistPDA's capability of understanding long-term spatiotemporal relationships in video streams under online inference settings, we introduce a novel STRD module, which can effectively transfer the spatiotemporal reasoning ability of existing VLMs from offline processing to online inference. To accommodate the tasks of online VAPDA, we construct a large-scale benchmark dataset, VAPDA-127K, which serves as a valuable resource for future research on online video anomaly understanding. Extensive experiments have shown that AssistPDA achieves superior performance compared to existing state-of-the-art VLMs across the VAP, VAD, and VAA tasks.

{
    \small
    \bibliographystyle{ieeenat_fullname}
    \bibliography{main}
}
\clearpage
\setcounter{page}{1}
% \maketitle
\maketitlesupplementary

\section{Dataset Construction Details}
\label{sec:intro}

This section provides additional details on the dataset construction process.

\subsection{Caption Model}
For generating video frame captions, we follow the work \cite{zanella2024harnessing-28} and aggregate the outputs from five caption models: BLIP2-flan-t5-xl, BLIP2-flan-t5-xl-coco, BLIP2-flan-t5-xxl, BLIP2-opt-6.7b, and BLIP2-opt-6.7b-coco. This aggregation helps mitigate potential biases from individual caption models.

\subsection{Task Prompts for LLM}
\begin{figure*}[t]
  \centering
   \includegraphics[width=1\linewidth]{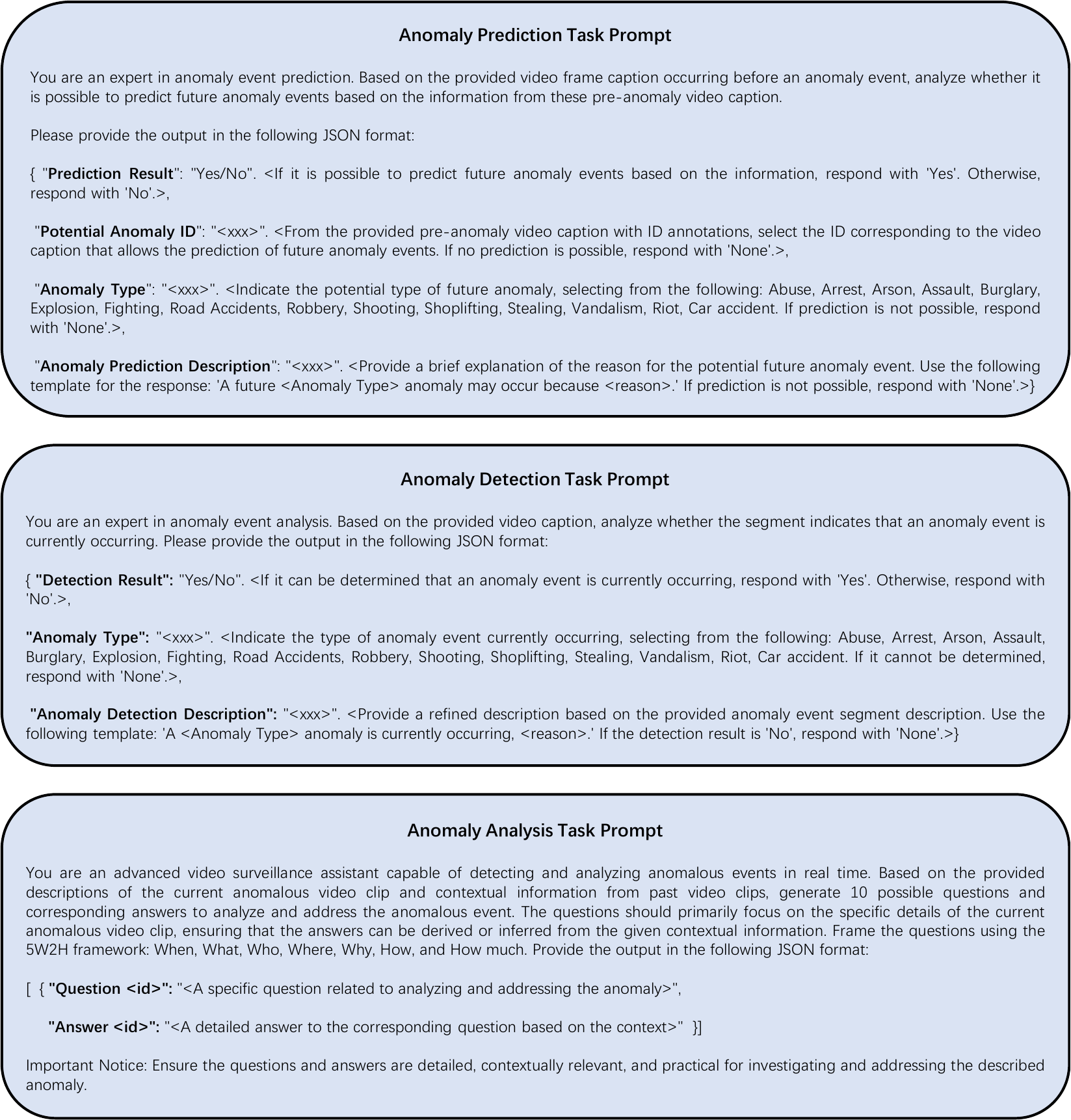}
   \caption{Illustration of how to prompt LLM to generate data for VAP, VAD, and VAA tasks.}
   \label{fig: prompt}
\end{figure*}

For the three tasks VAP, VAD, and VAA, task-specific data are generated based on captions using the Qwen2.5-72B-Instruct model, which is the most powerful open-source LLM available at the time of dataset construction. The specific task prompts for each task are shown in Figure \ref{fig: prompt}.

\vspace{-1.5mm}
\subsection{Dataset Splits}
\vspace{-1.5mm}
Table \ref{data} provides a detailed split of the training and test set composition of VAPDA-127K across the three tasks. Notably, for the anomaly analysis test set, we
select one of the ten open-ended questions corresponding to each timestamp as the final test sample to ensure compatibility with online streaming inference. 
% \vfill 
% \begin{center}
%   \includegraphics[width=\textwidth]{fig/Vis.pdf}
%   \captionof{figure}{Visualization results on the test set.}
%   \label{fig: vis}
% \end{center}
Table \ref{data_compare} compares our dataset with existing VAD-related datasets, highlighting its advantages. Our dataset provides textual annotations tailored for online video anomaly prediction and detection tasks. Additionally, for the anomaly analysis task, it offers open-ended question-answer pairs specific to individual anomalous events.
\vspace{-1.5mm}
\section{Impact of LoRA Fine-tuning Parameters}
\vspace{-1.5mm}
Table \ref{Table: LoRA} presents the impact of different LoRA fine-tuning parameters, \( r \) and \( \alpha \), on performance. The results show that different parameter combinations affect LM-PPL, TimeDiff, and Fluency metrics differently. After a comprehensive trade-off, we select \( r=32 \) and \( \alpha=64 \) as the optimal configuration.
\vspace{-1.5mm}
\section{More Qualitative Results}
\label{sec:intro}
Figure \ref{fig: vis} further demonstrates the qualitative results on the test set. For the VAP and VAD task, the assistant receives the video stream input in real time and gives a response at the moment when the anomaly may occur as well as at the moment when the anomaly actually occurs. For the VAA task, the assistant immediately responds to the user's question.

\vspace{-5mm}
\begin{strip}
  \centering
  \includegraphics[width=\textwidth]{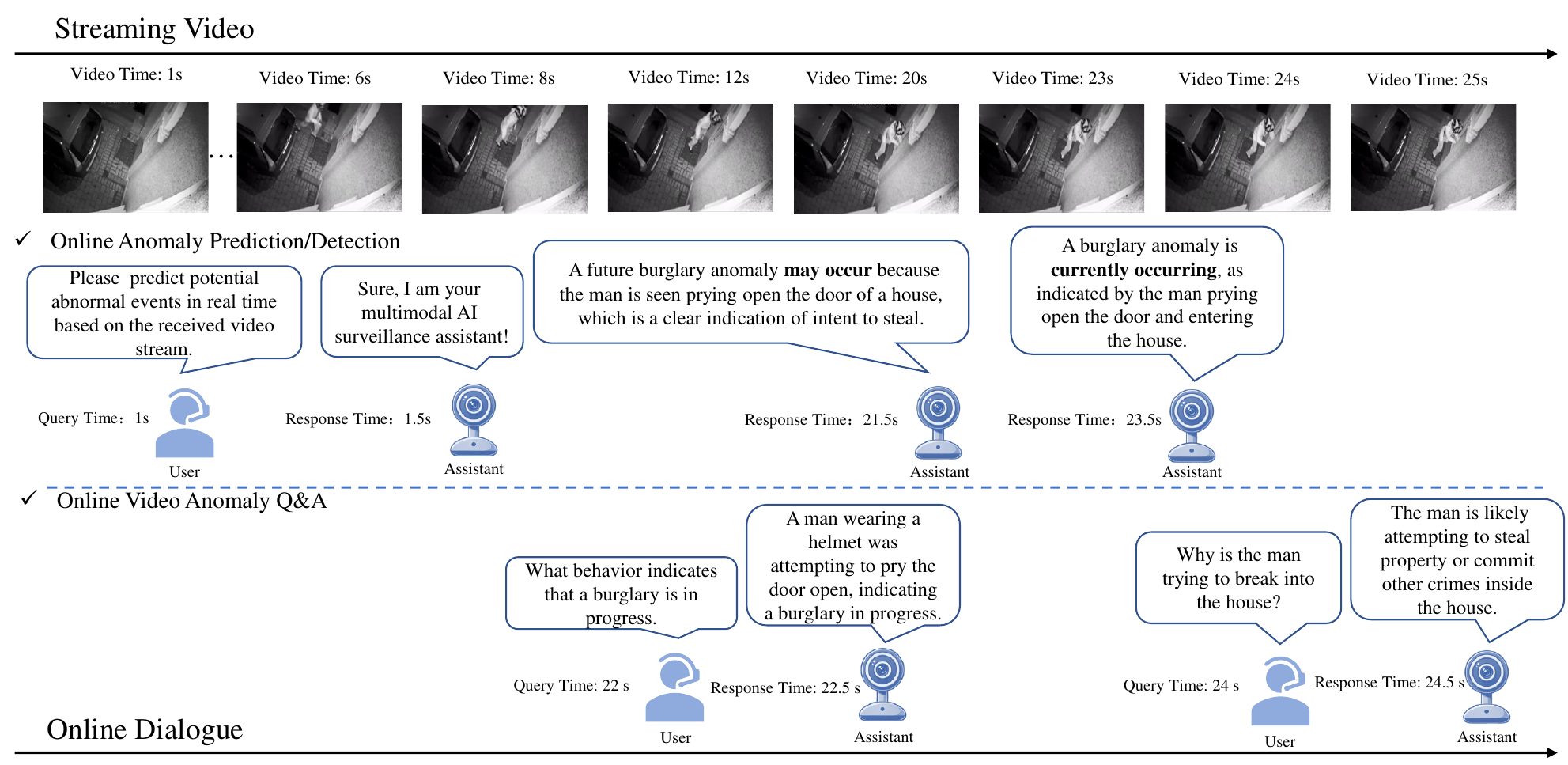}
  \captionof{figure}{Visualization results on the test set.}
  \label{fig: vis}
\end{strip}
% \vspace{22mm}
\vspace{-1.2em}

\begin{table*}
\centering
% \setlength{\tabcolsep}{1pt} 
% % \small
% % \footnotesize
% \scriptsize
\arrayrulecolor{black}
\begin{tabular}{lcccccc} 
\hline
~                                               & \multicolumn{3}{c}{VAP}      & \multicolumn{3}{c}{VAD}       \\ 
\cmidrule(lr){1-1}\cmidrule(lr){2-4}\cmidrule(lr){5-7}
~                                               & LM-PPL $\downarrow$ & TimeDiff $\downarrow$ & Fluency  $\uparrow$­ & LM-PPL $\downarrow$ & TimeDiff $\downarrow$ & Fluency  $\uparrow$­  \\ 
\hline
r=8/$\alpha$=16                                    & 1.69   & 1.09     & 53.41\%    & 1.98   & 4.72     & 46.55\%   \\
r=16/$\alpha$=32                                   & 1.67   & 1.07     & 53.74\%    & 1.99   & 4.68     & 46.76\%   \\
\rowcolor[rgb]{0.851,0.851,0.851} r=32/$\alpha$=64 & 1.68   & 1.07     & 53.81\% & 1.96   & 4.71     & 46.83\%   \\
r=64/$\alpha$=128                                  & 1.70   & 1.12     & 53.65\%  & 1.98   & 4.85     & 46.71\%   \\
\hline
\end{tabular}
\caption{Performance comparison of our method with different LoRA fine-tuning parameters.}
\label{Table: LoRA}
\end{table*}

\begin{table*}[t]
\centering
\begin{tabular}{ccccc} 
\hline
\multicolumn{5}{c}{\textbf{\textit{VAPDA-127K dataset}}}                                 \\ 
\hline
~           & Prediction text & Detection text & Anomaly Analysis (QA pair) & Timestamp  \\ 
\hline
Training set & 2511            & 6513           & 96720                      & \textcolor{green}{\checkmark}         \\
Test set    & 556             & 1521           & 19630 (1963)               & \textcolor{green}{\checkmark}          \\
\hline
\end{tabular}
\caption{Detailed training and test set split for VAPDA-127K dataset.}
\label{data}
\end{table*}

\begin{table*}[t]
\centering
\setlength{\tabcolsep}{5pt} 
% \small
% \footnotesize
\scriptsize
\begin{tabular}{ccccccccc} 
\hline
\multirow{3}{*}{Methods} & \multirow{3}{*}{\#Categories} & \multirow{3}{*}{\#Samples} & \multicolumn{4}{c}{Text}                                                                                            & \multirow{3}{*}{Temp. Anno.} & \multirow{3}{*}{VLM tuning}  \\ 
\cline{4-7}
                         &                               &                            & \multirow{2}{*}{Prediction text} & \multirow{2}{*}{Detection text} & \multicolumn{2}{c}{Anomaly Analysis (QA pair)} &                              &                              \\ 
\cline{6-7}
                         &                               &                            &                                  &                                 & Fixed template & Open-end                      &                              &                              \\ 
\hline
UCA \cite{yuan2024towards}                      & 13                            & 23542                      & \textcolor{red}{\ding{55}}                                & \textcolor{red}{\ding{55}}                               & \textcolor{red}{\ding{55}}              & \textcolor{red}{\ding{55}}                             & \textcolor{green}{\checkmark}                            & \textcolor{red}{\ding{55}}                            \\
LAVAD \cite{zanella2024harnessing-28}                   & N/A                           & N/A                        & \textcolor{red}{\ding{55}}                                & \textcolor{red}{\ding{55}}                               & \textcolor{red}{\ding{55}}              & \textcolor{red}{\ding{55}}                             & \textcolor{red}{\ding{55}}                            & \textcolor{red}{\ding{55}}                            \\
VAD-VideoLLama \cite{lv2024video-33}          & 13/7                          & 2400                       & \textcolor{red}{\ding{55}}                                & \textcolor{red}{\ding{55}}                               & \textcolor{red}{\ding{55}}              & \textcolor{red}{\ding{55}}                             & \textcolor{red}{\ding{55}}                            & projection                   \\
CUVA \cite{du2024uncovering-29}                    & 11                            & 6000                       & \textcolor{red}{\ding{55}}                               & \textcolor{red}{\ding{55}}                               & \textcolor{green}{\checkmark}              & \textcolor{red}{\ding{55}}                             & \textcolor{red}{\ding{55}}                            & \textcolor{red}{\ding{55}}                            \\
Hawk \cite{tang2024hawk-30}                    & -                             & 16000                      & \textcolor{red}{\ding{55}}                                & \textcolor{red}{\ding{55}}                              & \textcolor{green}{\checkmark}              & \textcolor{red}{\ding{55}}                             & \textcolor{red}{\ding{55}}                            & projection                   \\
HIVAU-70K \cite{zhang2024holmes-32}               & 19                            & 70000                      & \textcolor{red}{\ding{55}}                              & \textcolor{red}{\ding{55}}                               & \textcolor{green}{\checkmark}              & \textcolor{red}{\ding{55}}                             & \textcolor{green}{\checkmark}                            & LoRA                         \\
\rowcolor[rgb]{0.851,0.851,0.851} VAPDA-127K (Ours)        & 15                            & 127451                     & \textcolor{green}{\checkmark}                                & \textcolor{green}{\checkmark}                               & -              & \textcolor{green}{\checkmark}                            & \textcolor{green}{\checkmark}                       & LoRA                         \\
\hline
\end{tabular}
\caption{Comparison of other existing VAD method datasets.}
\label{data_compare}
\end{table*}

% \begin{figure*}
%   \centering
%   % \fbox{\rule{0pt}{2in} \rule{0.9\linewidth}{0pt}}
%    \includegraphics[width=1\linewidth]{fig/Vis.pdf}
%    \caption{further test set results are shown.}
%    \label{fig:vis}
% \end{figure*}

% During video preprocessing, Qwen2-VL supports naive dynamic resolutions and applies a smart resizing mechanism by setting a maximum pixel value. Here, we also adopt this functionality to maximize GPU resource utilization while avoiding GPU constraints imposed by processing long videos. These constraints would otherwise force an excessively low resolution for short videos, leading to an insufficient number of visual tokens and ultimately hindering visual understanding. This limitation would force the use of overly low resolutions, resulting in an insufficient number of tokens for short videos and ultimately hindering visual understanding.

%执行细节补充

\end{document}